\documentclass{article}

\usepackage[nonatbib, preprint]{neurips_2021}

\usepackage[utf8]{inputenc} 
\usepackage[T1]{fontenc}    
\usepackage{hyperref}       
\usepackage{url}            
\usepackage{booktabs}       
\usepackage{amsfonts}       
\usepackage{nicefrac}       
\usepackage{microtype}      
\usepackage{xcolor}         
\usepackage{bm}
\usepackage{amsmath}

\usepackage[pdftex]{graphicx} 
\usepackage{wrapfig}
\usepackage{lipsum}
\usepackage{multirow}
\usepackage{caption}
\usepackage{tabularx}
\usepackage{algorithm,algorithmicx,algpseudocode}
\usepackage{amssymb}
\usepackage{comment}

\title{Searching for Efficient Multi-Stage\\ Vision Transformers}

\author{%
  Yi-Lun Liao, 
  Sertac Karaman, 
  Vivienne Sze
  \\
  {Massachusetts Institute of Technology}
  \\
  \small{\texttt{\{ylliao, sertac, sze\}@mit.edu}}
}

\begin{document}

\maketitle

\newcommand{\todo}[1]{\textcolor{orange}{#1}}

\definecolor{crimson}{RGB}{163, 31, 52}

\begin{abstract}
\vspace{-1mm}
Vision Transformer (ViT) demonstrates that Transformer for natural language processing can be applied to computer vision tasks and result in comparable performance to convolutional neural networks (CNN), which have been studied and adopted in computer vision for years.
This naturally raises the question of how the performance of ViT can be advanced with design techniques of CNN.
To this end, we propose to incorporate two techniques and present ViT-ResNAS, an efficient multi-stage ViT architecture designed with neural architecture search (NAS).
First, we propose residual spatial reduction to decrease sequence lengths for deeper layers and utilize a multi-stage architecture.
When reducing lengths, we add skip connections to improve performance and stabilize training deeper networks.
Second, we propose weight-sharing NAS with multi-architectural sampling. 
We enlarge a network and utilize its sub-networks to define a search space.
A super-network covering all sub-networks is then trained for fast evaluation of their performance.
To efficiently train the super-network, we propose to sample and train multiple sub-networks with one forward-backward pass. 
After that, evolutionary search is performed to discover high-performance network architectures.
Experiments on ImageNet demonstrate that ViT-ResNAS achieves better accuracy-MACs and accuracy-throughput trade-offs than the original DeiT and other strong baselines of ViT.
Code is available at \url{https://github.com/yilunliao/vit-search}.

\end{abstract}
\vspace{-2mm}
\section{Introduction}
\vspace{-2mm}

Self-attention and Transformers~\cite{transformer}, which originated from natural language processing (NLP), have been widely adopted in computer vision (CV) tasks,  including image classification~\cite{aa_net, squeeze_and_excite, stand_alone_self_attention, botnet, non_local_network, exploring_self_attention}, object detection~\cite{detr, botnet, deformable_detr}, and semantic segmentation~\cite{max_deeplab, axial_deeplab}.
Many works utilize hybrid architectures and incorporate self-attention mechanisms into convolutional neural networks (CNN) to model long-range dependence and improve the performance of networks.
On the other hand, Vision Transformer (ViT)~\cite{vit} demonstrates that a pure transformer without convolution can achieve impressive performance on image classification when trained on large datasets like JFT-300M~\cite{jft_300m}.
Additionally, DeiT~\cite{deit} shows that ViT can outperform CNN when trained on ImageNet~\cite{imagenet} with stronger regularization.
It is appealing to have powerful Transformers for CV tasks since it enables using the same type of neural architecture for applications in both CV and NLP domains.

A parallel line of research is to design efficient neural networks with neural architecture search (NAS)~\cite{once_for_all, proxylessnas, spos, mobilenet_v3, single_path_automl, single_path_nas, mnasnet, hat, fbnet, netadapt, autoslim, bignas, nas_reinforcement, nasnet}.
Pioneering works use reinforcement learning to design efficient CNN architectures. 
They sample many networks in a pre-defined search space and train them from scratch for a few epochs to approximate their performance, which requires expensive computation.
To accelerate the process, weight-sharing NAS has become popular.
Instead of training individual networks in a search space, weight-sharing NAS trains a super-network whose weights are shared across all networks in the search space.
Once the super-network is trained, we can directly use its weights to approximate the performance of different networks in the search space.
These methods successfully result in CNN architectures outperforming manually designed ones.

While CNN architectures have been studied and adopted in CV for years and optimized with NAS, recently ViT demonstrates superior performance over CNN in some scenarios.
Despite its promising performance, ViT adopts the same architecture as Transformer for NLP~\cite{transformer}. 
This naturally leads to the question of how the performance of ViT can be further advanced with design techniques of CNN.
Therefore, in this work, we propose to incorporate two design techniques of CNN, which are spatial reduction and neural architecture search, and present ViT-ResNAS, an efficient multi-stage \textbf{ViT} architecture with \textbf{res}idual spatial reduction and designed with \textbf{NAS}.

\begin{figure}

   \includegraphics[width=0.9\linewidth]{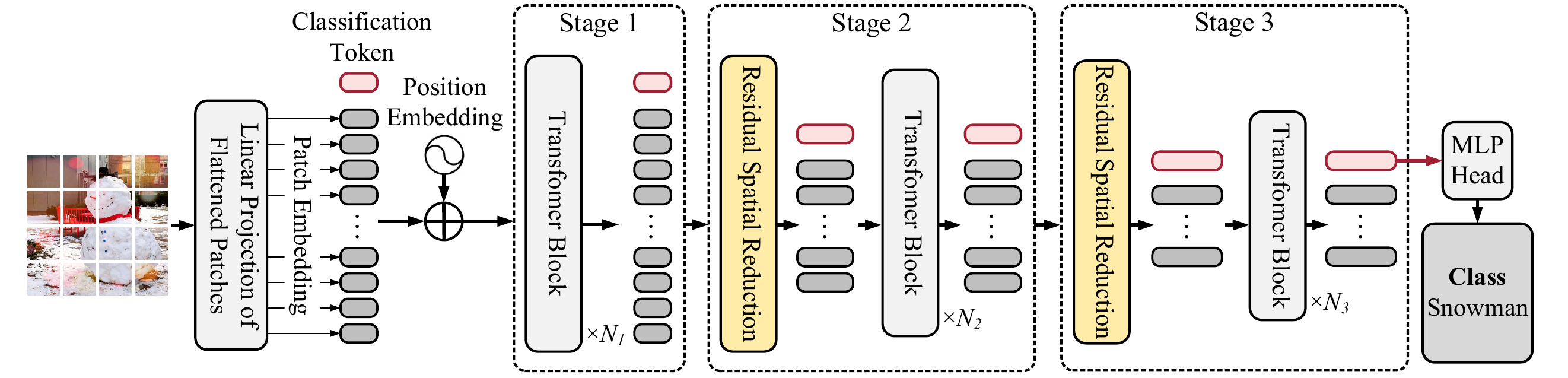}
   \centering
   \vspace{-1mm}
   \caption{\textbf{Architecture of ViT-Res.} We propose residual spatial reduction (light orange) to reduce sequence length and increase embedding size for deeper blocks, which divides the network into several stages. Each stage has the same sequence length and embedding size and consists of several transformer blocks. All stages except the first one start with residual spatial reduction blocks.}
   \vspace{-6mm}
\label{fig:vit_res}
\end{figure}

\begin{figure}[t]
    \centering
   \includegraphics[width=0.95\linewidth]{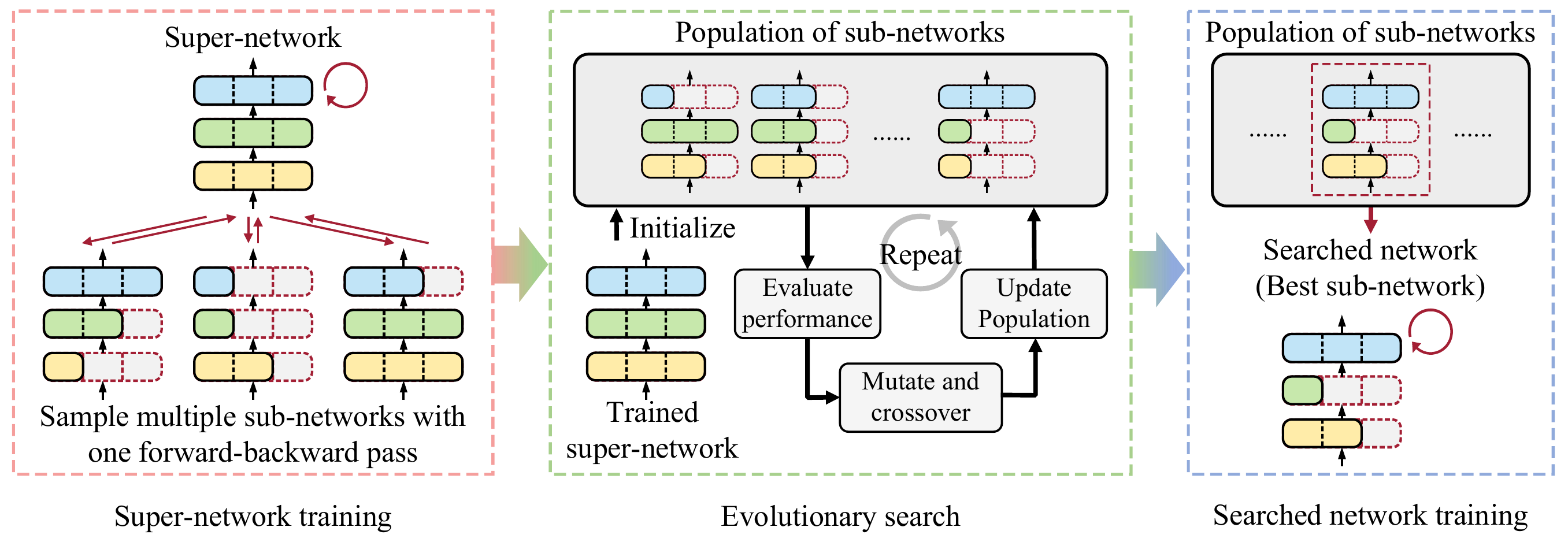}

   \vspace{-1mm}
   \caption{\textbf{Algorithm flow of NAS.}
   First, we train a ViT-Res super-network with multi-architectural sampling. The performance of sub-networks can be directly evaluated using the  super-network's trained weights without further training. Then, we perform evolutionary search to find high-performance sub-networks. Finally, the best sub-network becomes our searched network and is trained from scratch to convergence.}
   \vspace{-4mm}
\label{fig:nas_overview}
\end{figure}

First, we propose \textit{residual spatial reduction} to decrease sequence lengths and increase embedding sizes for deeper transformer blocks.
As illustrated in Fig.~\ref{fig:vit_res}, we transform the original single-stage architecture into a multi-stage one, with each stage having the same sequence length and embedding size.
This architecture follows that of CNN, where the resolution of feature maps decreases and the channel size increases for deeper layers.
Additionally, we add \textit{skip connections} when reducing sequence lengths, which can further improve performance and stabilize training deeper networks.
ViT with residual spatial reduction is named ViT-Res.
Second, we propose weight-sharing neural architecture search with \textit{multi-architectural sampling} to improve the architecture of ViT-Res as shown in Fig.~\ref{fig:nas_overview}.
We enlarge ViT-Res network by increasing its depth and width.
Its sub-networks are utilized to define a search space.
Then, a super-network covering all sub-networks is trained to directly evaluate their performance.
For each training iteration and given a batch of examples, we sample and train \textit{multiple} sub-networks with \textit{one} forward-backward pass to efficiently train the super-network.
Once the super-network is trained, evolutionary search~\cite{regularized_evolution, large_scale_evolution_classifier} is applied to discover high-performance ViT-ResNAS networks.
Experiments on ImageNet~\cite{imagenet} demonstrate the effectiveness of our proposed ViT-ResNAS. 
Compared to the original DeiT~\cite{deit}, ViT-ResNAS-Tiny achieves 8.6\% higher accuracy than DeiT-Ti with slightly higher multiply-accumulate operations (MACs), and ViT-ResNAS-Small achieves similar accuracy to DeiT-B while having $6.3 \times$ less MACs and $3.7 \times$ higher throughput.
Additionally, ViT-ResNAS achieves better accuracy-MACs and accuracy-throughput trade-offs than other strong baselines of ViT such as PVT~\cite{pvt} and PiT~\cite{pit}.

Our main contributions are as follows: 
(1) We propose residual spatial reduction to improve the efficiency of ViT.
(2) We propose weight-sharing NAS with multi-architectural sampling to improve ViT with residual spatial reduction (ViT-Res).
(3) Experiments on ImageNet demonstrate that our ViT-ResNAS achieves comparable accuracy-MAC trade-offs to previous works.

\vspace{-2mm}
\vspace{-1mm}
\section{Method}
\label{sec:method}


\vspace{-2mm}
We first review the architecture of Vision Transformer (ViT)~\cite{vit}. 
Then, we discuss residual spatial reduction and weight-sharing NAS with multi-architectural sampling to improve its architecture.
Other extra techniques that help improve performance are presented as well.

\vspace{-2mm}

\subsection{Background on Vision Transformer}
\vspace{-1mm}
Main components are tokenization, position embedding, multi-head self-attention (MHSA), feed-forward network (FFN), and layer normalization (LN). 
MHSA and FFN form a transformer block.

\vspace{-2.5mm}
\paragraph{Tokenization.}
The input to ViT is a fixed-size image and is split into \textit{patches} of pixels.
Each patch is transformed into a vector of dimension $d_{embed}$ called \textit{patch embedding} or \textit{patch token} with a linear layer.
Thus, an image is viewed as a sequence of patch embeddings. 
To perform classification, a learnable vector called classification token is appended to the sequence. 
After being processed with transformer blocks, the feature captured by the classification token is used to predict classes.

\vspace{-2.5mm}
\paragraph{Position Embedding.}
To preserve relative positions of image patches, position embeddings in the form of pre-defined or learnable vectors are added to patch embeddings.
Then, joint patch and position embeddings are processed with transformer blocks as shown on the left of Fig.~\ref{fig:vit_res}.

\vspace{-2.5mm}
\paragraph{MHSA.}
The attention function in MHSA operates on a sequence of embeddings $X \in \mathbb{R}^{N \times d_{embed}}$, with $N$ being sequence length and $d_{embed}$ being embedding dimension.
It first generates three vectors, \textit{key}, \textit{query} and \textit{value}, for each embedding with linear transformations, $W^K$, $W^Q$ and $W^V$, and packs them into matrices $K$, $Q$, and $V$ $\in \mathbb{R}^{N \times d}$.
Then, for each query corresponding to an embedding, we calculate its inner products with all $N$ keys, divide each product by $\sqrt{d}$, and apply softmax function over all products to obtain $N$ normalized weights.
The output of the attention function for the embedding is the weighted sum of $N$ value vectors.
The process is conducted for all embeddings and the attention function results in an output as follows:
\begin{equation}
\begin{aligned}
\text{Attn}(X, W^K, W^Q, W^V) = \text{softmax}(\frac{QK^T}{\sqrt{d}})V 
\text{, where } K = XW^K, Q = XW^Q, V = XW^V
\end{aligned}
\end{equation}
One MHSA contains $h$ parallel attention functions or attention heads.
The $h$ different outputs are concatenated and projected with a linear transformation to produce the output of MHSA.

\vspace{-2.5mm}
\paragraph{FFN.}
It consists of two linear layers separated by GeLU~\cite{gelu}. 
The first layer expands the dimension of each embedding to hidden size $d_{hidden}$ and the second projects the dimension back to $d_{embed}$.

\vspace{-2.5mm}
\paragraph{LN.} Layer normalization~\cite{layer_norm} normalizes each embedding $x$ in the sequence $X$ separately as follows: 
\begin{equation}
\text{LN}(x) = \frac{x - \mu}{\sigma} \circ \gamma + \beta
\end{equation}
$\mu$ and $\sigma$ are the mean and standard deviation of $x$ and the calculation is \textit{batch-independent}. 

\vspace{-2mm}
\subsection{Residual Spatial Reduction}
\label{subsec:residual_spatial_reduction}
\vspace{-1mm}

\begin{wrapfigure}[19]{R}{0.49\textwidth}
\vspace{-8pt}
  \begin{center}
    \includegraphics[width=0.49\textwidth]{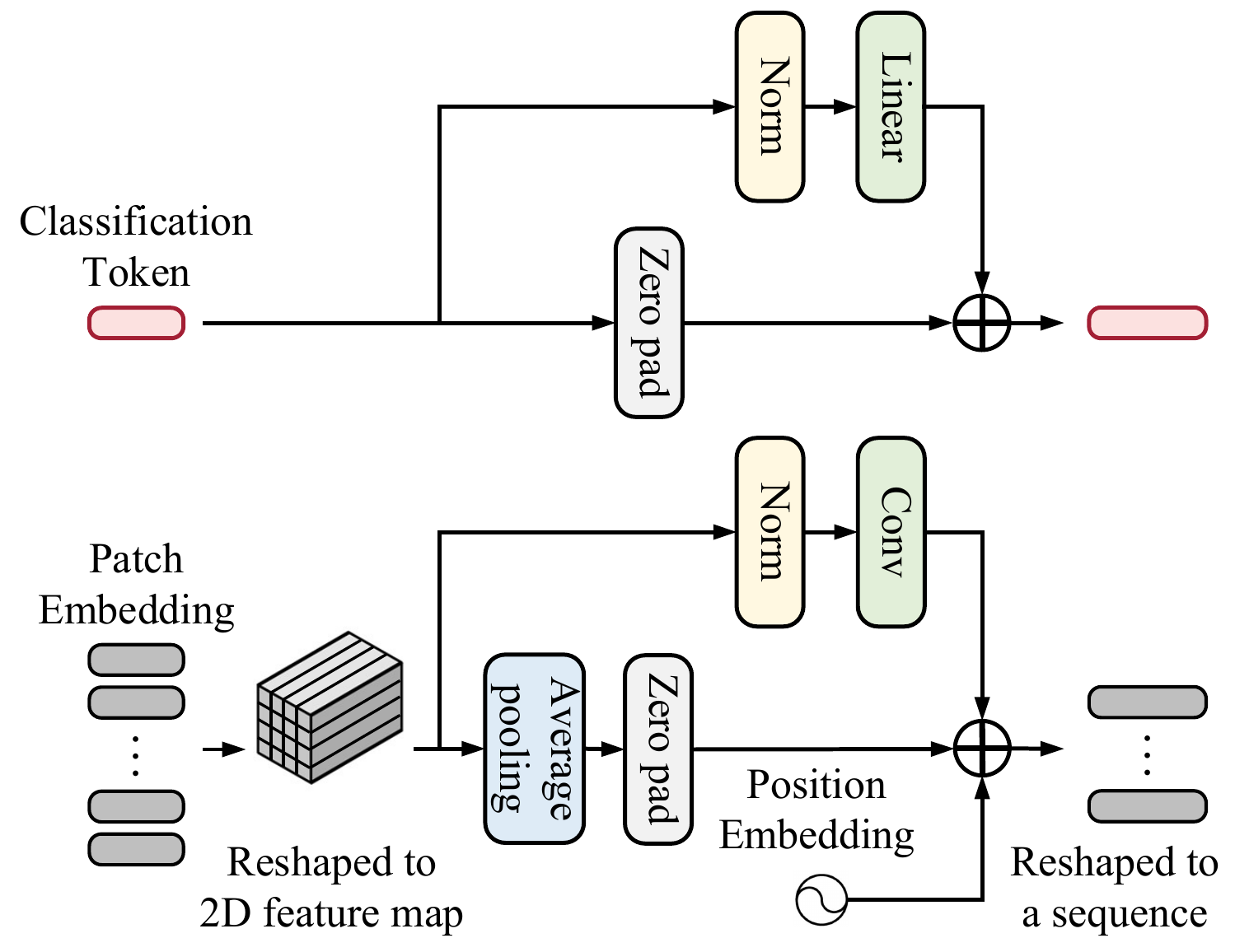}
  \end{center}
  \vspace{-9pt}
  \caption{\textbf{Structure of residual spatial reduction.} We use strided convolution to reduce sequence length of patch embeddings. A skip connection as shown in the lower branch is added to stabilize training and improve performance.}
  \label{fig:residual_spatial_reduction}
\end{wrapfigure}

ViT maintains the same sequence length throughout the network, which is similar to maintaining the same resolution of feature maps.
In contrast, CNNs decrease the resolution of feature maps and increase the channel size for deeper layers. 
The former is based on the fact that there is spatial redundancy in feature maps, and the latter is to assign more channels to high-level features in deeper layers and to maintain similar computation when resolution is decreased.
The design technique has been widely adopted in high-performance CNNs~\cite{resnet, regnet}, which motivates whether it can be introduced to improve the efficiency of ViT as well.
To this end, we propose residual spatial reduction.
As illustrated in Fig.~\ref{fig:residual_spatial_reduction}, since there is spatial relationship in patch embeddings, we reshape the 1D sequence into a 2D feature map and then apply layer normalization~\cite{layer_norm} and strided convolution (``Norm'' and ``Conv'' in Fig.~\ref{fig:residual_spatial_reduction}) to reduce the sequence length of patch embeddings and increase embedding dimension.
Since the sequence length is changed, we update the relative position information by adding new \textit{position embeddings}.
To maintain the same channel size of all embeddings, we apply layer normalization and a linear layer (``Norm'' and ``Linear'' in Fig.~\ref{fig:residual_spatial_reduction}) to the classification token.
These constitute the \textit{residual branch}.

Although introducing only the residual branch can significantly improve the accuracy-MAC trade-offs, training deeper networks can be unstable.
Specifically, under the training setting of DeiT-distill~\cite{deit}, using the residual branch significantly improves the accuracy of DeiT-Tiny from 74.5\% to 79.6\% with little increase in MACs.
However, training loss can become ``NaN'' when training deeper networks like our ViT-Res super-networks.
To remedy this, we introduce an extra skip connection~\cite{resnet} without any learnable operations as motivated by the residual structure of transformer blocks~\cite{transformer, layer_norm_in_transformer}.
We use \textit{2D average pooling} to reduce sequence length (``Average pooling'' in Fig.~\ref{fig:residual_spatial_reduction}) and concatenate embeddings with zero tensors (``Zero pad'' in Fig.~\ref{fig:residual_spatial_reduction}) to increase embedding dimension.
The structure results in the \textit{main branch} and helps to stabilize training and improve performance.

\begin{wrapfigure}[15]{R}{0.4\textwidth}
\vspace{-12pt}
  \begin{center}
    \includegraphics[width=0.4\textwidth]{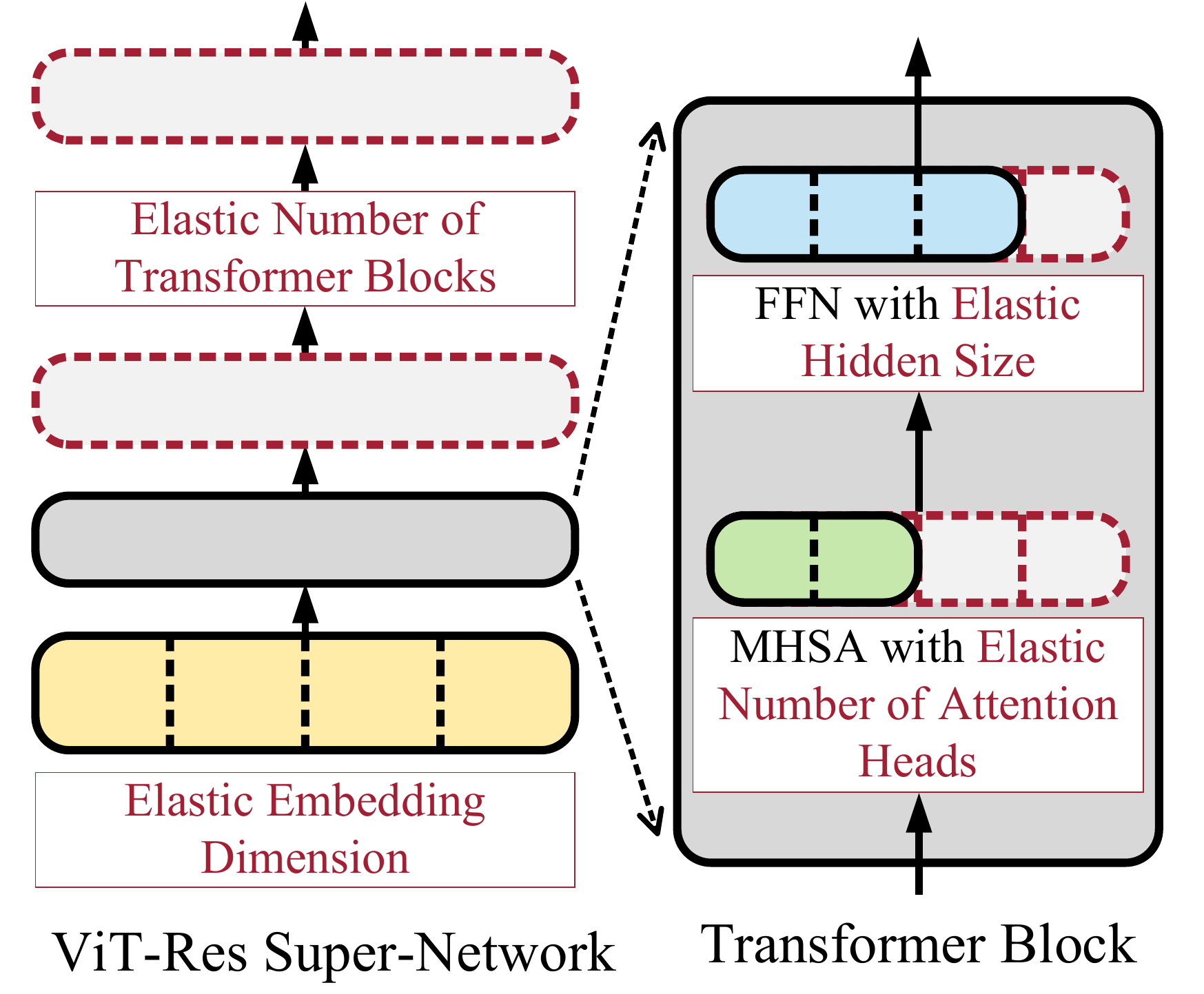}
  \end{center}
  \vspace{-1em}
  \caption{\textbf{Searchable dimensions in each stage of ViT-Res super-network.} 
  }
  \label{fig:supernet_search_space}
\end{wrapfigure}

The residual and main branches form \textit{residual spatial reduction}.
We insert 2 residual spatial reduction blocks into ViT with 12 blocks and divide the architecture evenly into 3 stages.
Following the design rule of ResNet~\cite{resnet}, we double embedding dimension when halving spatial resolution or reducing sequence length of patch embeddings by $4 \times$. 
The resulting ViT architecture is called ViT-Res as illustrated in Fig.~\ref{fig:vit_res}.
Since the spatial resolution is to be reduced by $4\times$ and input images are of size $224 \times 224$, the patch size in tokenization before the first stage is set to $14$, resulting in spatial resolution $16$ ($=\frac{224}{14}$) and $16 \times 16$ patch embeddings in the first stage.
After passing through two residual spatial reduction blocks, the sequence length of patch embeddings is reduced to $4 \times 4$.
Please refer to Appendix~\ref{appendix:vit_res_architecture} for further details.

\vspace{-2mm}
\subsection{Weight-Sharing NAS with Multi-Architectural Sampling}
\vspace{-1mm}

Another design technique used in CNNs is neural architecture search (NAS).
Due to its efficiency, we adopt weight-sharing NAS to improve the archtiecture of ViT-Res in terms of designing better \textit{architectural parameters} like numbers of attention heads and transformer blocks as shown in Fig.~\ref{fig:supernet_search_space}.

\vspace{-2.5mm}
\paragraph{Algorithm Overview.}
We enlarge ViT-Res network by increasing its depth and width.
Sub-networks with smaller depths and widths define a search space.
Then, a super-network covering all sub-networks is trained to directly evaluate their performance.
For each training iteration and given a batch of examples, we propose to sample \textit{multiple different} sub-networks with \textit{one} forward-backward pass to efficiently train the super-network.
After that, evolutionary search is applied to discover architectures of high-performance sub-networks satisfying some resource constraints like MACs.
Finally, the best sub-network evaluated becomes the searched network and is trained from scratch.

\vspace{-2.5mm}
\paragraph{Search Space.}
We construct a large network by \textit{uniformly} increasing depth and width of all stages of ViT-Res, and sub-networks contained in the large network define a search space.
As shown in Fig.~\ref{fig:supernet_search_space}, for \textit{each stage}, we search \textit{embedding dimension} and the \textit{number of transformer blocks}.
For \textit{different blocks}, we search \textit{different numbers of attention heads} $h$ in MHSA and \textit{different hidden sizes} $d_{hidden}$ in FFN.
The range of each \textit{searchable dimension} is pre-defined. 
During super-network training and evolutionary search, sub-networks of different configurations are sampled for training and evaluation.
Details of search space are presented in Appendix~\ref{appendix:search_space}.

\begin{figure}
\centering
   \includegraphics[width=0.75\linewidth]{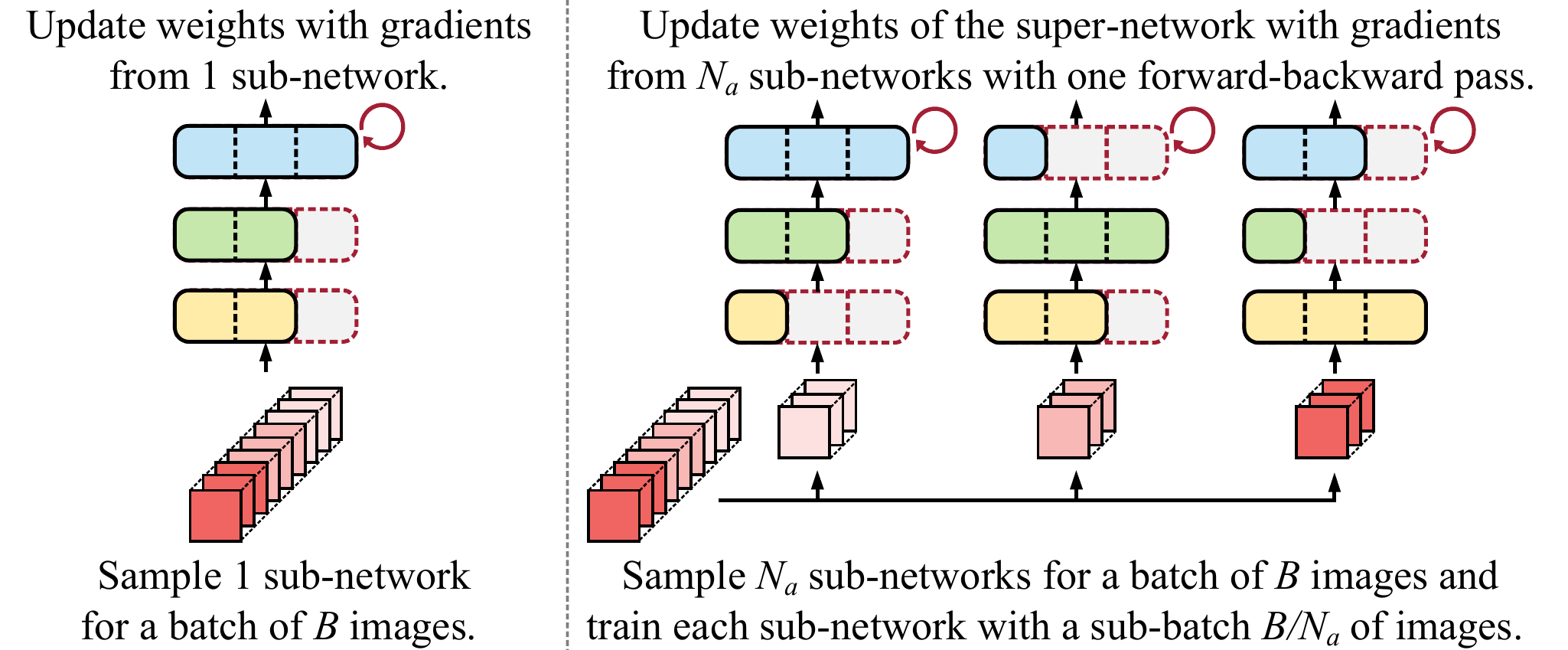}
   \vspace{-1mm}
   \caption{\textbf{Multi-architectural sampling for super-network training.} \textbf{Left:} one sub-network is sampled and trained with one forward-backward pass given a batch of examples. \textbf{Right:} multi-architectural sampling samples $N_a$ ($=3$) different sub-networks with one forward-backward pass.}
   \vspace{-6mm}
\label{fig:multi_architectural_sampling}
\end{figure}

\vspace{-2.5mm}
\paragraph{Multi-Architectural Sampling for Super-Network Training.}
To evaluate the performance of sub-networks, their weights have to be optimized to reflect their relative quality.
Therefore, we train a super-network whose architecture is the same as the largest network defining the search space and whose weights are shared across sub-networks.
During super-network training, we sample and train different sub-networks for different training iterations.
Generally, the more sub-networks are sampled, the more accurate the relative quality of sub-networks indicated by the trained super-network can be. 
Previous works on NAS for CNNs~\cite{once_for_all, spos, bignas} sample and train a \textit{single} sub-network with \textit{one} forward-backward pass for each training iteration.
Given a fixed amount of iterations, this, however, results in room for improvement when training ViT-Res super-networks as we can sample \textit{multiple} sub-networks with \textit{one} forward-backward pass for each training iteration as illustrated in Fig.~\ref{fig:multi_architectural_sampling}.

\begin{figure}
\centering
   \includegraphics[width=0.9\linewidth]{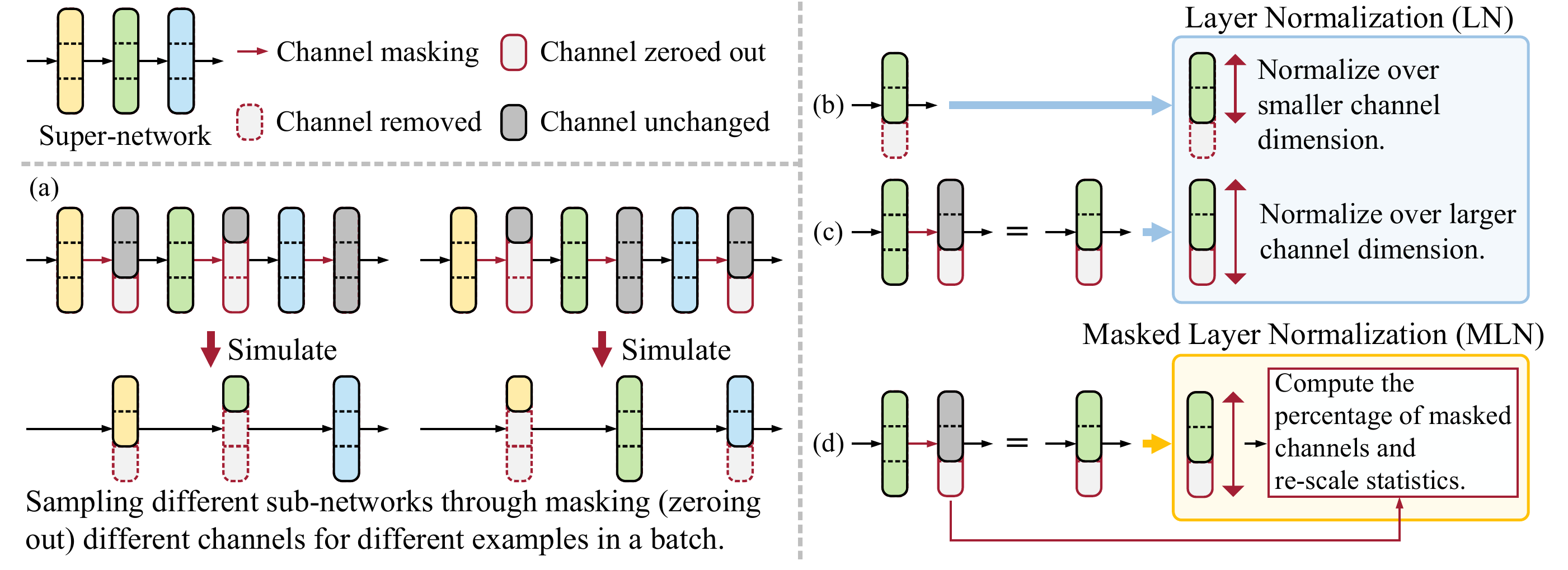}

   \vspace{-1mm}
   \caption{\textbf{Channel masking and masked layer normalization for multi-architectural sampling.} \textbf{(a)} We simulate sampling different sub-networks by masking different channels. \textbf{(b)} The channel dimension LN normalizes over when training a sub-network alone. \textbf{(c)} The channel dimension LN incorrectly normalizes over when sampling a sub-network during super-network training. \textbf{(d)} MLN re-calibrates statistics by considering the number of masked channels.
   }
   \vspace{-5mm}
\label{fig:implementation_multi_architectural_sampling}
\end{figure}

Unlike batch normalization~\cite{bn} in CNNs, layer normalization (LN)~\cite{layer_norm} in ViT avoids normalizing along batch dimension, which enables sampling \textit{different} sub-networks with \textit{one} forward-backward pass without mixing their statistics.
Specifically, for each training iteration, we sample $N_a$ sub-networks and divide a batch into $N_a$ sub-batches.
Each sub-network is trained with its corresponding sub-batch.
This can be achieved efficiently with \textit{one} forward-backward pass and \textit{channel masking} (\textit{ordered dropout})~\cite{fbnetv2, netadaptv2, bignas}.
As shown in Fig.~\ref{fig:implementation_multi_architectural_sampling} (a), different masks are generated for different feature maps corresponding to different examples in order to \textit{zero out} different channels and simulate sampling different sub-networks.
The shapes of feature maps remain the same, which maintains regular batch computation and therefore enables a single forward-backward pass.

Additionally, we \textit{re-calibrate} the statistics in LN when sampling multiple sub-networks in order to prevent discrepancy between super-network training and standard training.
As shown in Fig.~\ref{fig:implementation_multi_architectural_sampling} (b) and (c), LN can incorrectly normalize over a larger channel dimension when sampling networks with smaller channel sizes.
This is because channel masking only changes the values of feature maps not shapes.
To avoid the problem, we propose \textit{masked layer normalization} (MLN), which re-calibrates the statistics of LN when its input is masked.
Instead of only looking at the shape of input tensors, MLN calculates the ratio of the number of masked channels to the total number of channels and uses that ratio to re-calibrate statistics as in Fig.~\ref{fig:implementation_multi_architectural_sampling} (d).
With MLN, the statistics become the same as we train sub-networks separately as in Fig.~\ref{fig:implementation_multi_architectural_sampling} (b).
Eventually, with channel masking and MLN, we can sample \textit{multiple} sub-networks with \textit{one} forward-backward pass, which improves sample efficiency when training ViT-Res super-networks and therefore the performance of searched networks.
Other details of super-network training can be found in Appendix~\ref{appendix:additional_training_details}.

\vspace{-2.5mm}
\paragraph{Evolutionary Search.}
Throughout the search process, we only sample sub-networks satisfying pre-defined resource constraints (e.g., MACs) and evaluate their performance (e.g., accuracy) with trained super-network weights.
Evolutionary search maintains a population of networks and refines top-performing ones for many iterations.
We start with an initial population of $P_0$
randomly sampled sub-networks.
At every iteration, $N_{parent}$ sub-networks with the highest performance in the population serve as parent networks that generate $N_{child}$ new sub-networks through \textit{mutation} and \textit{crossover}.
For mutation, we randomly select one sub-network from parent networks and modify every architectural parameter with a pre-defined probability $p_{mutate}$.
For crossover, we choose two random sub-networks from parent networks and randomly fuse every architectural parameter.
Mutation and crossover generates the same amount of new sub-networks, and they are added to the population.
The process is repeated for $T_{search}$ times, and the best sub-network in the population becomes the searched network and is trained from scratch to convergence.

\vspace{-3mm}
\subsection{Extra Techniques}
\label{subsec:extra_techniques}
\vspace{-1mm}
We present other techniques in training and architecture that help improve performance below.

\vspace{-2.5mm}
\paragraph{Token Labeling with CutMix and Mixup.}
We incorporate \textit{token labeling}~\cite{vit_oversmooth, token_labeling} and propose to improve it with Mixup~\cite{mixup}.
Token labeling provides labels for all patches in an input image.
This enables training ViT to predict class labels of all \textit{patch tokens} (patch embeddings) in addition to predicting class label of an input image with classification token and can improve training Transformers~\cite{electra, vit_oversmooth, token_labeling}.
Token labeling generates an input image, its class label, and patch-wise class labels by patch-wise CutMix~\cite{cutmix} described as follows.
Given two images $x_1, x_2$ and their class labels $y_1, y_2$, we divide each image into $K$ patches, flatten the patches, and associate each patch with its original image-level class label.
In this way, we have sequences of flattened patches $X_1 = [x_1^1, ..., x_1^K]$ and $X_2 = [x_2^1, ..., x_2^K]$ and sequences of patch-wise labels $Y_1 = [y_1, ..., y_1]$ and $Y_2 = [y_2, ..., y_2]$.
Then, we randomly generate a binary mask $M$ to combine the images and associated patch-wise labels:
$X = X_1 \odot M + X_2 \odot (1 - M)$ and  
$Y = Y_1 \odot M + Y_2 \odot (1 - M)$.
We restructure the 1D sequence $X$ into a 2D image $x$.
The class label $y$ associated with $x$ is obtained by combining the original class labels:
$y = \lambda y_1 + (1 - \lambda) y_2$
where $\lambda$ is the average of elements in the binary mask $M$.
For ViT-Res, we choose $K = 4 \times 4$, which is equal to the sequence length of patch embeddings in the \textit{last} stage.

Additionally, we find that applying Mixup~\cite{mixup} along with token labeling can improve performance, which is contrary to previous results~\cite{vit_oversmooth, token_labeling}.
We surmise that they first apply Mixup and then perform patch-wise CutMix could lead to noisy training images and labels.
In contrast, we improve token labeling with Mixup through \textit{switching} between patch-wise CutMix and Mixup.
Specifically, we choose either patch-wise CutMix or Mixup to generate an image, its label and patch-wise labels.
When the former is selected, we follow the practice as mentioned above.
When the latter is chosen, the image and its label are generated in the same way as Mixup, and this image-level label is assigned to all patches to produce the patch-wise labels.

\vspace{-2.5mm}
\paragraph{Convolution before Tokenization.}
Following previous works~\cite{vit, levit, token_labeling}, we add some convolutional layers before tokenization.
Specifically, we add three extra $3 \times 3$ convolutional layers, each with $C$ output channels.
The stride of the first layer is set to $2$, and others are $1$.
A residual skip connection is added between outputs of the first and the third layers.
The computation of these convolutional layers is relatively low.
Since the spatial resolution is reduced by $2 \times$, the patch size in tokenization is set to $7$ ($= \frac{14}{2}$).
Further details can be found in Appendix~\ref{appendix:vit_res_architecture} and~\ref{appendix:search_space}.
\vspace{-2mm}
\section{Experiments}
\vspace{-2mm}

\definecolor{mygreen}{RGB}{64, 179, 79}

In this section, we first describe the experiment setup.
Then, we conduct experiments to study the effectiveness of our proposed methods.
Finally, the comparison with related works is presented.

\vspace{-2mm}
\subsection{Experiment Setup}
\vspace{-1mm}
The dataset used is ImageNet~\cite{imagenet}.
Our implementation is based on timm library~\cite{timm} and that of DeiT~\cite{deit}.
Most of the training settings follow those in DeiT~\cite{deit} except that we do not use repeated augmentation~\cite{repeated_augment}.
We train models with $8$ GPUs for $300$ epochs, and batch size is $1024$ and input resolution is $224 \times 224$.
As for token labeling, we adopt the same loss function as previous works~\cite{vit_oversmooth, token_labeling}, and the associated loss is directly added to the original classification loss without any scaling.
Unless otherwise stated, we incorporate token labeling with patch-wise CutMix and Mixup to train our networks and include several convolutional layers before tokenization.
For ablation study presented in Section~\ref{subsec:ablation_study}, when token labeling is used, drop path~\cite{stochastic_depth} rate is increased from $0.1$ to $0.2$. 

For neural architecture search, we increase the depth and width of our ViT-Res network to build search spaces and search the architectures of ViT-ResNAS networks of three different sizes, which we name Tiny, Small, and Medium.
For ViT-ResNAS-Tiny, we enlarge ViT-Res-Tiny to build a super-network and train both the super-network and the searched network with drop path rate $0.2$.
For ViT-ResNAS-Small and Medium, we further enlarge ViT-Res-Tiny super-network and share the same search space.
The super-network is trained with drop path rate $0.3$.
The drop path rates for Small and Medium are $0.3$ and $0.4$, respectively.
The details of the search space are presented in Table~\ref{appendix:search_space} in appendix.
A super-network is trained for $120$ epochs, with other settings being the same as mentioned above.
We sample $N_a$ sub-networks with one forward-backward pass during super-network training, experiment with different values of $N_a$ and empirically set $N_a$ to $16$.
For evolutionary search, the resource constraint is MAC.
We set search iteration $T_{search} = 20$, the number of parent networks $N_{parent} = 75$, the initial population size $P_0 = 500$, the number of new sub-networks for each iteration $N_{child} = 150$ and mutation probability $p_{mutate} = 0.3$.


In addition, following DeiT~\cite{deit}, we fine-tune networks at larger resolutions to obtain higher capacity.
A network is fine-tuned for $30$ epochs, with batch size $512$, learning rate $5 \times 10^{-6}$, weight decay $10^{-8}$ and drop path rate $0.75$.

\vspace{-3mm}
\begin{table}
\centering
\scalebox{0.9}{
\begin{tabular}{l|l}
\toprule[1.2pt]
Methods                                  & Top-1 Acc.  \\ 
\midrule[1.2pt]
Spatial reduction & 78.5 \\
Residual spatial reduction & 78.8 \textcolor{mygreen}{(+0.3)} \\ 
\midrule
+ Token labeling with CutMix & 79.6 \textcolor{mygreen}{(+0.8)} \\ 
+ Token labeling with CutMix \& Mixup & 80.1 \textcolor{mygreen}{(+0.5)} \\
\bottomrule[1.2pt]
\end{tabular}
}
\vspace{2mm}
\caption{\textbf{Additive study of improving multi-stage network with residual connections and token labeling.} We start with DeiT-Tiny with spatial reduction and progressively introduce residual connections and token labeling. Adding residual connections and combining token labeling with Mixup in the proposed manner improve accuracy without increasing MACs.}
\vspace{-4mm}
\label{tab:residual_connection_token_labeling}
\end{table}


\subsection{Ablation Study}
\label{subsec:ablation_study}

\vspace{-2mm}
\paragraph{Multi-Stage Network with Residual Connection and Token Labeling.}
We study how the performance of vanilla multi-stage networks can be enhanced with the proposed residual connections and improved token labeling training.
We build such a network by starting with DeiT-Tiny network, adding three convolutional layers before tokenization and inserting two spatial reduction blocks (i.e., only the \textit{residual branch} of residual spatial reduction).
The results are shown in Table.~\ref{tab:residual_connection_token_labeling}.
Without token labeling, introducing only two residual connections (i.e., \textit{main branch} of residual spatial reduction) can improve accuracy from $78.5\%$ to $78.8\%$ with negligible overhead.
When token labeling is used, incorporating Mixup in the proposed manner can further improve the accuracy by $0.5\%$.
With residual spatial reduction and token labeling, we can improve the accuracy by $1.6\%$ without increasing MACs, and our ViT-Res-Tiny achieves $80.1\%$ top-1 accuracy with $1.8$G MACs. 


We also study the effectiveness of residual connections in the proposed residual spatial reduction under another training setting.
When using the training setting of DeiT-distill~\cite{deit}, which includes repeated augmentation and distilling knowledge of CNN, residual connections can improve the accuracy from $79.6\%$ to $80.1\%$.
Moreover, when we train deeper networks such as our ViT-Res-Tiny super-network, without residual connections, the training can be unstable with ``NaN'' training loss.

\begin{table}
\centering
\scalebox{0.9}
{
\begin{tabular}{l|c|c|c|c}
\toprule[1.2pt]
\multirow{2}{*}{\begin{tabular}[c]{@{}l@{}}Number of sampled sub-networks\\ per forward-backward pass ($N_a$) \end{tabular}} & Single-arch.              & \multicolumn{3}{c}{Multi-arch.}                  \\ \cmidrule{2-5} 
                                                                                              & \multicolumn{1}{c|}{1} & \multicolumn{1}{c|}{8} & \multicolumn{1}{c|}{16} & \multicolumn{1}{c}{32} \\ 
\midrule
Top-1 accuracy (\%) & 80.5 &  \begin{tabular}[c]{@{}c@{}}80.6\\ \textcolor{mygreen}{(+0.1)}\end{tabular} & \begin{tabular}[c]{@{}c@{}}80.8\\ \textcolor{mygreen}{(+0.3)}\end{tabular} & \begin{tabular}[c]{@{}c@{}}80.6\\ \textcolor{mygreen}{(+0.1)}\end{tabular} \\
\bottomrule[1.2pt]
\end{tabular}
}
\vspace{2mm}
\caption{\textbf{Effect of numbers of sampled sub-networks with one forward-backward pass} ($N_a$) \textbf{on the performance of searched networks.} 
The type of resource constraint is MAC, and all searched networks have MAC around $1.8$G. 
Empirically, $N_a = 16$ results in the best searched network.
}
\vspace{-8mm}
\label{tab:multi_architectural_sampling}
\end{table}

\vspace{-2.5mm}
\paragraph{Weight-Sharing NAS with Multi-Architectural Sampling.} 
With the proposed residual spatial reduction and token labeling, we study how weight-sharing NAS with multi-architectural sampling can further improve the performance of ViT-Res.
During super-network training, we sample and train $N_a$ sub-networks with one forward-backward pass. 
Given the same amount of training iterations and training examples, the value of $N_a$ controls the trade-offs between sample efficiency (i.e., how many sub-networks are sampled) and the quality of training each sub-network (i.e., how many examples are used to train it).
Therefore, instead of arbitrarily choosing a large value, we experiment with different $N_a$.
We use our ViT-Res-Tiny super-network to study the effect of different $N_a$ on the performance of searched networks and search for networks with MACs around $1.8$G.
The results are presented in Table.~\ref{tab:multi_architectural_sampling}.
Compared to sampling one sub-network for each training iteration (``Single-arch.'' in Table~\ref{tab:multi_architectural_sampling}), sampling multiple sub-networks leads to better searched networks.
Among different values, $N_a = 16$ empirically results in the best searched network. 
Please note that with NAS, the top-1 accuracy is increased from $80.1\%$ to $80.5\%$ and that with the proposed multi-architectural sampling, the accuracy is further increased from $80.5\%$ to $80.8\%$.
Based on the results, we choose $N_a = 16$ to design larger ViT-ResNAS networks. 

\vspace{-2mm}
\subsection{Comparison with Related Works}
\vspace{-1mm}

\begin{table}
\centering
\scalebox{0.8}{
\begin{tabular}{l|c|c|c|c}
\toprule[1.2pt]
Method          & Model Size (M)& MACs (G)&
\begin{tabular}[c]{@{}c@{}}Top-1\\ Accuracy (\%) \end{tabular} & 
\begin{tabular}[c]{@{}c@{}}Throughput\\ (images/second) \end{tabular} \\
\midrule[1.2pt]
DeiT-Ti~\cite{deit} & 5 & 1.3     & 72.2 & 1968 \\
T2T-ViT-12 ~\cite{t2t_vit} & 7 & 2.2     & 76.5 & 1192 \\
PiT-XS~\cite{pit} & 11 & 1.4     & 78.1 & 1647 \\
PVT-Tiny~\cite{pvt} & 13 & 1.9     & 75.1 & 1133 \\
ViL-Tiny~\cite{vil} & 7 & 1.3     & 76.3 & 754 \\
\textbf{ViT-Res-Tiny} &  43  & 1.8     & 80.1 & 1807 \\
\textbf{ViT-ResNAS-Tiny} & 41 & 1.8     & 80.8 & 1579 \\

\midrule
DeiT-Small~\cite{deit} & 22     & 4.6     & 79.9 & 846 \\
T2T-ViT-14~\cite{t2t_vit} & 22 & 5.2     & 81.5 & 682 \\
PiT-S~\cite{pit} & 24 & 2.9     & 80.9 & 986\\
PVT-Small~\cite{pvt} & 25 & 3.8     & 79.8 & 628 \\
PVT-Medium~\cite{pvt} & 44 & 6.7 & 81.2 & 407 \\
TNT-S~\cite{tnt} & 24 & 5.2     & 81.5 & 353 \\
Swin-T~\cite{swin} & 29 & 4.5     & 81.3 & 600 \\
Twins-PCPVT-S~\cite{twins} & 24 & 3.7 & 81.2 & 622 \\
\textbf{ViT-ResNAS-Small} & 65 & 2.8     & 81.7 & 1084 \\
\midrule
DeiT-Base~\cite{deit} & 86 & 17.6    & 81.8 & 290 \\
T2T-ViT-19~\cite{t2t_vit} & 39 & 8.9     & 81.9 & 428 \\
PiT-B~\cite{pit} & 74 & 12.5    & 82.0 & 316 \\
PVT-Large~\cite{pvt} & 61 & 9.8     & 81.7 & 284 \\
ViL-Small~\cite{vil} & 25 & 4.9 & 82.0 & 310 \\
CvT-13~\cite{cvt} & 20 & 4.5 & 81.6 & 587 \\
\textbf{ViT-ResNAS-Medium} & 97 & 4.5 & 82.4 & 751 \\

\midrule
Swin-S~\cite{swin} & 50 & 8.7 & 83.0 & 351 \\
Twins-PCPVT-B~\cite{twins} & 44 & 6.4 & 82.7 & 403 \\
CvT-21~\cite{cvt} & 32 & 7.1 & 82.5 & 379 \\
\textbf{ViT-ResNAS-Medium $\uparrow 280$} & 97 & 7.1 & 83.1 & 467 \\

\midrule
ViL-Medium-Wide~\cite{vil} & 40 & 11.3 & 82.9 & 177 \\
Twins-PCPVT-L~\cite{twins} & 61 & 9.5 & 83.1 & 282 \\
CaiT-S36~\cite{cait} & 68 & 13.9 & 83.3 & 191 \\
\textbf{ViT-ResNAS-Medium $\uparrow 336$} & 97 & 10.6 & 83.5 & 292 \\

\midrule
Swin-B~\cite{swin} & 88 & 15.4 & 83.3 & 243 \\
CaiT-XS24 $\uparrow 384$~\cite{cait} & 27 & 19.3 & 83.8 & 57 \\
\textbf{ViT-ResNAS-Medium $\uparrow 392$} & 97 & 15.2 & 83.8 & 194\\

\bottomrule[1.2pt]
\end{tabular}
}
\vspace{2mm}
\caption{\textbf{Comparison with related works on ViT.} ``$\uparrow R$'' denotes that the model is first trained at resolution $224$ and then fine-tuned at resolution $R$. Other models are trained at resolution $224$. Throughput is measured on one Titan RTX GPU with batch size $128$.}
\vspace{-8mm}
\label{tab:comparison}
\end{table}

We design our ViT-ResNAS-Tiny, Small and Medium networks with NAS and MACs as our search metric.
Following DeiT~\cite{deit}, we fine-tune our ViT-ResNAS-Medium at larger resolutions to obtain models with higher capacity.
Since the patch size before the first stage is $14$ and there are $2$ residual spatial reduction in the network, the spatial resolution is reduced by $56$ $(=14 \times 2 \times 2)$ times in the last stage, and therefore we can only increase the input resolution by multiples of $56$.
We also report the performance of fine-tuning ViT-ResNAS-Medium at resolutions $280$, $336$ and $392$.
When comparing different models, following the implementation of Swin Transformer~\cite{swin}, we measure the inference throughput on a single Titan RTX GPU with batch size 128 as well.

Table~\ref{tab:comparison} summarizes the comparison with previous works on ViT. 
Please note that why our models have more parameters is that we further reduce the sequence length and increase the channel size of DeiT models. 
Even though we have more parameters, our ViT-ResNAS networks consistently achieves better accuracy-MACs trade-offs as well as accuracy-throughput trade-offs.

Compared to the original single-stage DeiT~\cite{deit}, ViT-ResNAS-Tiny achieves $8.6\%$ higher accuracy than DeiT-Ti while having only $0.5$G higher MACs and slightly lower throughput and has $0.9\%$ higher accuracy than DeiT-Small with $2.5 \times$ less MACs and $1.9 \times$ higher throughput.
ViT-ResNAS-Small obtains the similar level of accuracy to DeiT-Base while having $6.3 \times$ lower MACs and $3.7 \times$ higher throughput.
Compared to other works on multi-stage architectures like PVT~\cite{pvt} and PiT~\cite{pit}, ViT-ResNAS consistently has better accuracy-MACs and accuracy-throughput trade-offs.
The computation saving becomes more apparent as accuracy becomes higher.
For example, in comparison to PiT-S~\cite{pit}, ViT-ResNAS-Tiny achieves similar accuracy while having $1.6 \times$ less MACs and $1.6 \times$ higher throughput.
When compared with PiT-B~\cite{pit}, ViT-ResNAS-Medium achieves $0.4 \%$ higher accuracy with $2.8 \times$ less MACs and $ 2.4 \times$ higher throughput.
This suggests the effectiveness of NAS to scale up models. 

Additionally, for lower MACs around 2.0G, ViT-ResNAS-Tiny achieves better accuracy-MACs trade-offs than ViL-Tiny.
However, when MACs are around $5.0$G, ViT-ResNAS-Medium is on par with ViL-Small in terms of accuracy-MACs trade-offs.
This probably suggests that utilizing efficient attention mechanisms~\cite{twins, swin, vil} with \textit{convolution-like locality} to process high-resolution features is necessary for models in higher MAC regimes to generalize better.
Nevertheless, those methods are orthogonal to our approaches, and the proposed weight-sharing NAS with multi-architectural sampling could further improve their accuracy-MACs trade-offs as well.

\vspace{-2mm}
\section{Related Works}
\label{sec:related_work}
\vspace{-2mm}

\paragraph{Vision Transformers.}
Vision Transformer (ViT)~\cite{vit} demonstrates that a pure transformer without convolution can perform well on image classification when trained on large datasets like JFT-300M~\cite{jft_300m}.
To make it data-efficient, DeiT~\cite{deit} uses strong regularization and adds a distillation token to its architecture for knowledge distillation, and demonstrates comparable performance when trained on ImageNet~\cite{imagenet} only.
Subsequent works improve the performance of ViT on ImageNet through either better training~\cite{vit_oversmooth, token_labeling} or architectures~\cite{convit, tnt, pit, swin, hvt, cait, pvt, cvt, t2t_vit, vil}.
For example, they bring locality into network architectures by using convolutions~\cite{pvt, cvt} or efficient local attention~\cite{twins, tnt, swin, vil} or similarly adopt multi-stage architectures~\cite{pit, hvt, pvt, cvt, vil}.
Our proposed methods are complementary to these works.
Residual spatial reduction can derive a more efficient multi-stage architectures, and weight-sharing NAS with multi-architectural sampling can be applied to further improve performance since they use layer normalization~\cite{layer_norm} as well.

\vspace{-2.5mm}
\paragraph{Neural Architecture Search.}
There have been increasingly interest in designing efficient architectures with neural architecture search (NAS)~\cite{tunas, once_for_all, proxylessnas, spos, mobilenet_v3, bossnas, regularized_evolution, large_scale_evolution_classifier, evolved_transformer, single_path_automl, single_path_nas, mnasnet, e3d, hat, fbnet, netadapt, netadaptv2, autoslim, bignas, nas_reinforcement, nasnet}.
Among different methods, weight-sharing NAS~\cite{tunas, once_for_all, proxylessnas, spos, bossnas, darts, enas,  single_path_automl, single_path_nas, hat, fbnet, autoslim, bignas} has become popular due to efficiency.
They train one over-parametrized super-network whose weights are shared across all networks in a search space to conduct architecture search, which saves computation cost significantly.
However, many of these works focus on CNN, which has been researched for years, and have well-designed search space.
In contrast, our proposed NAS with multi-architectural sampling focuses on multi-stage ViT, which is much less studied, and we utilize its batch-independent property to further improve the performance of searched networks.

\vspace{-2mm}
\section{Conclusion}
\vspace{-2mm}
We incorporate two design techniques of CNN, which are spatial reduction and NAS, into ViT and present ViT-ResNAS, an efficient multi-stage ViT designed with NAS.
The proposed residual spatial reduction enhances accuracy-MACs trade-offs significantly and the residual connections can improve performance and stabilize training.
Weight-sharing NAS with multi-architectural sampling trains super-networks more efficiently and results in better searched networks.
Experiments on ImageNet demonstrates the effectiveness of our methods and the efficiency of our ViT-ResNAS networks.



\newpage
{\small
\bibliographystyle{plain}
\bibliography{ref}}


\newpage
\appendix
\section*{Appendix}

\section{Network Architecture of ViT-Res} 
\label{appendix:vit_res_architecture}
We build our ViT-Res-Tiny network by introducing two modifications to DeiT-Tiny~\cite{deit}.
First, residual spatial reduction (RSR) is applied to evenly divide the single-stage architecture into a multi-stage one as described in Sec.~\ref{subsec:residual_spatial_reduction}.
For ``Conv'' in Fig.~\ref{fig:residual_spatial_reduction}, we use $3 \times 3$ convolutions to downsample patch embeddings.
Second, we add three convolutional layers as mentioned in Sec.~\ref{subsec:extra_techniques} and use $C = 24$.
Table~\ref{tab:appendix:vit_res_architecture} summarizes the archtiecture of ViT-Res-Tiny.

\begin{table}[h!]
\centering
\scalebox{0.9}{
\begin{tabular}{l|c|c}
\toprule[1.2pt]
                              & \multicolumn{1}{l|}{\begin{tabular}[c]{@{}l@{}}Output Sequence\\ Length\end{tabular}} & \textbf{ViT-Res-Tiny}                                                   \\ 
\midrule[1.2pt]
\multirow{4}{*}{Tokenization} & \multirow{4}{*}{$16 \times 16 + 1$}                                                          & conv-3\_24\_2                                                  \\ 
& & $\left[ \begin{tabular}[c]{@{}c@{}}conv-3\_24\_1\\ conv-3\_24\_1\end{tabular}\right]$ \\
                              &                                                                                       & conv-7\_192\_7                                                 \\ 
\midrule
Stage 1                       & $16 \times 16 + 1$                                                                           & $\left[ \begin{tabular}[c]{@{}c@{}}MHSA-64\_3\\ FFN-768\end{tabular}\right] \times 4$   \\ 
\midrule[1pt]
\multirow{3}{*}{Stage 2}      & $8 \times 8 + 1$                                                                             & RSR-384                                                        \\ \cmidrule{2-3} 
                              & $8 \times 8 + 1$                                                                        & $\left[\begin{tabular}[c]{@{}c@{}}MHSA-64\_6\\ FFN-1536\end{tabular}\right] \times 4$   \\ 
\midrule[1pt]
\multirow{3}{*}{Stage 3}      & $4\times4+1$                                                                            & RSR-768                                                        \\ \cmidrule{2-3} 
                              & $4\times4+1$                                                                             & $\left[\begin{tabular}[c]{@{}c@{}}MHSA-64\_12\\ FFN-3072\end{tabular}\right] \times 4$ \\ 
\bottomrule[1.2pt]
\end{tabular}
}
\vspace{2mm}
\caption{\textbf{Architecture of ViT-Res-Tiny.} ``conv-$k$\_$C$\_$s$'' stands for a $k \times k$ convolutional layer with output channel $C$ and stride $s$. ``MHSA-$d$\_$h$'' is MHSA with head dimension $d$ and $h$ attention heads. ``FFN-$d_{hidden}$'' is FFN with hidden size $d_{hidden}$. ``RSR-$d_{embed}$'' is residual spatial reduction with output embedding dimension $d_{embed}$. Note that the embedding 
dimension in Stage 1 is determined by the last convolutional layer in tokenization.}
\label{tab:appendix:vit_res_architecture}
\end{table}

\section{Search Space}
\label{appendix:search_space}
We enlarge ViT-Res-Tiny to build ViT-Res-Tiny super-network.
We increase numbers of attention heads $h$ and decrease head dimensions $d$ for the first two stage so that we have more choices of attention heads $h$.
We do not search configurations of convolutional layers in tokenization except that we search the number of output channels in the last layer, which determines the embedding dimension of the first stage.
For the search space for ViT-ResNAS-Tiny, each stage has three pairs of transformer blocks.
The first block in each pair always remains while the second one is skippable and can be removed.
Therefore, each stage can have three to six transformer blocks.
Additionally, we further enlarge the ViT-Res-Tiny super-network to construct ViT-Res-Small and Medium super-network and search space by increasing width and adding one extra block for each stage.
Table~\ref{tab:appendix:search_space} summarizes the search spaces for ViT-ResNAS networks.

\begin{figure}
    \begin{minipage}{0.5\linewidth}
\centering
\scalebox{0.8}{
\begin{tabular}{l|c|c}
\toprule[1.2pt]
                              & \multicolumn{1}{l|}{\begin{tabular}[c]{@{}l@{}}Output \\ Length\end{tabular}} & \begin{tabular}[c]{@{}c@{}}\textbf{ViT-ResNAS-Tiny}\\ \textbf{Search Space} \end{tabular}                                       \\ 
\midrule[1.2pt]
\multirow{5}{*}{Token.} & \multicolumn{2}{l}{$d_1 \in \{256, 224, 192, 176, 160\}$}                                                                                                                                                    \\ \cmidrule{2-3} 
                              & \multirow{4}{*}{$16 \times 16 + 1$}                                                          & conv-3\_24\_2                                                                                                \\ 
& & $\left[ \begin{tabular}[c]{@{}c@{}}conv-3\_24\_1\\ conv-3\_24\_1\end{tabular}\right]$ \\
                              &                                                                                       & conv-7\_$d_1$\_1\_7                                                                                              \\ 
\midrule
\multirow{8}{*}{Stage 1}      & \multicolumn{2}{l}{\begin{tabular}[c]{@{}l@{}}$h_1 \in \{6, 5, 4, 3\}$, \\ $f_1 \in \{768, 704, 640, 576, 512, 448, 384\}$\end{tabular}}                                                                           \\ \cmidrule{2-3} 
                              & $16 \times 16 + 1 $                                                                          & $\left[ \begin{tabular}[c]{@{}c@{}}
                              $\left(
                                \begin{tabular}[c]{@{}c@{}}
                                    MHSA-32\_$h_1$\\ 
                                    FFN-$f_1$\\
                                \end{tabular} \right)$ \\
                              \\ 
                              
                              $\left(
                                \begin{tabular}[c]{@{}c@{}}
                                    (skippable)\\ 
                                    MHSA-32\_$h_1$\\ 
                                    FFN-$f_1$\\
                                \end{tabular} \right)$ \\
                              \end{tabular} \right] \times 3$  \\ 
\midrule[1pt]
\multirow{10}{*}{Stage 2}      & \multicolumn{2}{l}{\begin{tabular}[c]{@{}l@{}}$d_2 \in \{512, 448, 384, 352, 320\}$, \\ $h_2 \in \{12, 10, 8, 6\}$, \\ $f_2 \in \{1536, 1408, 1280, 1152, 1024,$ \\ $896, 768\}$ \end{tabular}}                                          \\ \cmidrule{2-3} 
                              & $8 \times 8 + 1$                                                                             & RSR-$d_2$                                                                                                     \\ \cmidrule{2-3} 
                              & $8 \times 8 + 1$                                                                             & $\left[ \begin{tabular}[c]{@{}c@{}}
                              $\left(
                                \begin{tabular}[c]{@{}c@{}}
                                    MHSA-48\_$h_2$\\ 
                                    FFN-$f_2$\\
                                \end{tabular} \right)$ \\
                              \\ 
                              
                              $\left(
                                \begin{tabular}[c]{@{}c@{}}
                                    (skippable)\\ 
                                    MHSA-48\_$h_2$\\ 
                                    FFN-$f_2$\\
                                \end{tabular} \right)$ \\
                              \end{tabular} \right] \times 3$ \\ 
\midrule[1pt]
\multirow{10}{*}{Stage 3}      & \multicolumn{2}{l}{\begin{tabular}[c]{@{}l@{}}$d_3 \in \{1024, 896, 768, 704, 640\}$, \\ $h_3 \in \{12, 10, 8, 6\}$, \\ $f_3 \in \{3072, 2816, 2560, 2304, 2048,$ \\$1792, 1536\}$ \end{tabular}}                                          \\ \cmidrule{2-3} 
                              & $4 \times 4 + 1$                                                                             & RSR-$d_3$                                                                                                     \\ \cmidrule{2-3} 
                              & $4 \times 4 + 1$                                                                             & $\left[ \begin{tabular}[c]{@{}c@{}}
                              $\left(
                                \begin{tabular}[c]{@{}c@{}}
                                    MHSA-64\_$h_3$\\ 
                                    FFN-$f_3$\\
                                \end{tabular} \right)$ \\
                              \\ 
                              
                              $\left(
                                \begin{tabular}[c]{@{}c@{}}
                                    (skippable)\\ 
                                    MHSA-64\_$h_3$\\ 
                                    FFN-$f_3$\\
                                \end{tabular} \right)$ \\
                              \end{tabular} \right] \times 3$ \\ 
\bottomrule[1.2pt]
\end{tabular}
}
\end{minipage}
\hfill
\begin{minipage}{0.5\linewidth}
\centering
\scalebox{0.8}{
\begin{tabular}{l|c|c}
\toprule[1.2pt]
                              & \multicolumn{1}{l|}{\begin{tabular}[c]{@{}l@{}}Output \\ Length\end{tabular}} & \begin{tabular}[c]{@{}c@{}}\textbf{ViT-ResNAS-Small and}\\ \textbf{Medium Search Space} \end{tabular}                                       \\ 
\midrule[1.2pt]
\multirow{5}{*}{Token.} & \multicolumn{2}{l}{$d_1 \in \{320, 280, 240, 220, 200\}$}                                                                                                                                                    \\ \cmidrule{2-3} 
                              & \multirow{4}{*}{$16 \times 16 + 1$}                                                          & conv-3\_24\_2                                                                                                \\ 
& & $\left[ \begin{tabular}[c]{@{}c@{}}conv-3\_24\_1\\ conv-3\_24\_1\end{tabular}\right]$ \\
                              &                                                                                       & conv-7\_$d_1$\_1\_7                                                                                              \\ 
\midrule
\multirow{8}{*}{Stage 1}      & \multicolumn{2}{l}{\begin{tabular}[c]{@{}l@{}}$h_1 \in \{8, 7, 6, 5\}$, \\ $f_1 \in \{960, 880, 800, 720, 640, 560, 480\}$\end{tabular}}                                                                           \\ \cmidrule{2-3} 
                              & $16 \times 16 + 1 $                                                                          & 
                              \begin{tabular}[c]{@{}c@{}}
                                $\left[ \begin{tabular}[c]{@{}c@{}}
                                    $\left(
                                    \begin{tabular}[c]{@{}c@{}}
                                    MHSA-32\_$h_1$\\ 
                                    FFN-$f_1$\\
                                    \end{tabular} \right)$ \\
                                \\ 
                                
                                $\left(
                                    \begin{tabular}[c]{@{}c@{}}
                                    (skippable)\\ 
                                    MHSA-32\_$h_1$\\ 
                                    FFN-$f_1$\\
                                    \end{tabular} \right)$ \\
                                \end{tabular} \right] \times 3$  
                                \\
                                \\
                              
                                $\left(
                                    \begin{tabular}[c]{@{}c@{}}
                                    MHSA-32\_$h_1$\\ 
                                    FFN-$f_1$\\
                                    \end{tabular} \right)$ 
                            \end{tabular}
                              \\ 
\midrule[1pt]
\multirow{10}{*}{Stage 2}      & \multicolumn{2}{l}{\begin{tabular}[c]{@{}l@{}}$d_2 \in \{640, 560, 480, 440, 400\}$, \\ $h_2 \in \{16, 14, 12, 10\}$, \\ $f_2 \in \{1920, 1760, 1600, 1440, 1280,$ \\ $1120, 960\}$ \end{tabular}}                                          \\ \cmidrule{2-3} 
                              & $8 \times 8 + 1$                                                                             & RSR-$d_2$                                                                                                     \\ \cmidrule{2-3} 
                              & $8 \times 8 + 1$                                                                             & 
                              \begin{tabular}[c]{@{}c@{}}
                                $\left[ \begin{tabular}[c]{@{}c@{}}
                                    $\left(
                                    \begin{tabular}[c]{@{}c@{}}
                                    MHSA-48\_$h_2$\\ 
                                    FFN-$f_2$\\
                                    \end{tabular} \right)$ \\
                                \\ 
                                
                                $\left(
                                    \begin{tabular}[c]{@{}c@{}}
                                    (skippable)\\ 
                                    MHSA-48\_$h_2$\\ 
                                    FFN-$f_2$\\
                                    \end{tabular} \right)$ \\
                                \end{tabular} \right] \times 3$  
                                \\
                                \\
                              
                                $\left(
                                    \begin{tabular}[c]{@{}c@{}}
                                    MHSA-48\_$h_2$\\ 
                                    FFN-$f_2$\\
                                    \end{tabular} \right)$ 
                            \end{tabular}
                            \\ 
\midrule[1pt]
\multirow{10}{*}{Stage 3}      & \multicolumn{2}{l}{\begin{tabular}[c]{@{}l@{}}$d_3 \in \{1280, 1120, 960, 880, 800\}$, \\ $h_3 \in \{16, 14, 12, 10\}$, \\ $f_3 \in \{3840, 3520, 3200, 2880, 2560, $ \\$2240, 1920\}$ \end{tabular}}                                          \\ \cmidrule{2-3} 
                              & $4 \times 4 + 1$                                                                             & RSR-$d_3$                                                                                                     \\ \cmidrule{2-3} 
                              & $4 \times 4 + 1$                                                                             & \begin{tabular}[c]{@{}c@{}}
                                $\left[ \begin{tabular}[c]{@{}c@{}}
                                    $\left(
                                    \begin{tabular}[c]{@{}c@{}}
                                    MHSA-64\_$h_3$\\ 
                                    FFN-$f_3$\\
                                    \end{tabular} \right)$ \\
                                \\ 
                                
                                $\left(
                                    \begin{tabular}[c]{@{}c@{}}
                                    (skippable)\\ 
                                    MHSA-64\_$h_3$\\ 
                                    FFN-$f_3$\\
                                    \end{tabular} \right)$ \\
                                \end{tabular} \right] \times 3$  
                                \\
                                \\
                                $\left(
                                    \begin{tabular}[c]{@{}c@{}}
                                    MHSA-64\_$h_3$\\ 
                                    FFN-$f_3$\\
                                    \end{tabular} \right)$ 
                            \end{tabular} \\ 
\bottomrule[1.2pt]
\end{tabular}
}
\end{minipage}
\vspace{2mm}
\captionof{table}{\textbf{ViT-ResNAS search space.}
\textbf{Left:} search space for ViT-ResNAS-Tiny.
\textbf{Right:} search space for ViT-ResNAS-Small and Medium.
For each stage $i$, we search embedding dimension $d_{embed}$ and numbers of transformer blocks.
For each transformer block, we search the number of attention heads $h$ in MHSA and hidden size $d_{hidden}$ in FFN.
Different blocks can have different values of $h$ and $d_{hidden}$.
The first rows in tokenization and each stage define the range of each searchable dimension, with $d_i$, $h_i$, and $f_i$ corresponding to $d_{embed}$, $h$, and $d_{hidden}$ in stage $i$, respectively.
Transformer blocks with ``skippable'' can be removed during super-network training and evolutionary search, which supports different numbers of blocks in searched networks.
}
\vspace{10mm}
\label{tab:appendix:search_space}
\end{figure}

\section{Additional Training Details}
\label{appendix:additional_training_details}
\paragraph{Super-Network Training.}
We divide the original training set into \textit{sub-train} and \textit{sub-validation} sets.
The sub-validation set contains $25$K images, with 25 images for each class.
The rest of images form the sub-train set.
We train super-networks on the sub-train set, and during evolutionary search, we evaluate the accuracy of sub-networks on the sub-validation set.

Additionally, during super-network training, we \textit{warm up} training different widths and depths (filter warmup~\cite{tunas, netadaptv2} or progressive shrinking~\cite{once_for_all, e3d}). 
Specifically, at the beginning, we only sample and train the largest sub-network in a search space.
As the super-network training proceeds, we \textit{gradually} sample and train sub-networks with smaller widths and depths.
After 25\% of total training epochs, sub-networks with any width and depth can be sampled and trained.

\paragraph{Training Cost.}
We report the time for training networks and performing NAS when 8 V100 (16GB) GPUs are used.
Training ViT-Res-Tiny takes about 32 hours.
For ViT-ResNAS-Tiny, super-network training takes about 16.7 hours, evolutionary search takes 5.5 hours, and it takes about 34.5 hours to train the searched ViT-ResNAS-Tiny.
For ViT-ResNAS-Small and Medium, the cost of training the shared super-network is 21 hours.
For ViT-ResNAS-Small, the evolutionary search takes 6 hours and training the searched network takes 41.6 hours.
For ViT-ResNAS-Medium, it takes 6 hours and 45 hours to perform evolutionary search and training the searched network, respectively.
Note that we report the training time just for completeness and that the time can vary across different platforms.

\section{Searched ViT-ResNAS Architecture}
\label{appendix:searched_archtiecture}
We report the searched architectures of ViT-ResNAS-Tiny, Small and Medium in Table~\ref{tab:appendix:vit_resnas_tiny},~\ref{tab:appendix:vit_resnas_small}, and~\ref{tab:appendix:vit_resnas_medium}, respectively.
Compared to ViT-Res-Tiny, ViT-ResNAS-Tiny has smaller embedding dimensions but more transformer blocks.
From ViT-ResNAS-Tiny to Small, stage 1 has one extra block, and embedding dimensions are increased uniformly.
As for ViT-ResNAS-Medium, the first two stages have more blocks than the last stage.
These suggest that having more blocks in earlier stages is more important to scale up networks.


\section{Limitation}
\label{appendix:limitation}
We discuss some limitations of our approaches.
First, the performance of searched networks designed with NAS relies on manually designed search spaces.
We build our search space by uniformly increasing widths and depths of ViT-Res-Tiny, which is designed with some simple heuristics and without tuning.
Compared to search spaces of CNN (e.g., MobileNet), our search space is less studied and less optimized.
Optimizing search spaces such as setting a better range for each architectural parameter could potentially result in additional performance gain.
Second, our searched networks use the same type of attention mechanism as ViT, which could have large memory consumption when processing high-resolution feature maps (long sequences).
Incorporating more efficient attention mechanisms~\cite{twins, swin, vil} into search spaces could solve the issue and result in networks with better performance.

\begin{table}[ht]
\centering
\scalebox{0.9}{
\begin{tabular}{l|c|c}
\toprule[1.2pt]
                              & \multicolumn{1}{l|}{\begin{tabular}[c]{@{}l@{}}Output\\ Length\end{tabular}} & \textbf{ViT-ResNAS-Tiny}                                                   \\ 
\midrule[1.2pt]
\multirow{4}{*}{Tokenization} & \multirow{4}{*}{$16 \times 16 + 1$}                                                          & conv-3\_24\_2                                                  \\ 
& & $\left[ \begin{tabular}[c]{@{}c@{}}conv-3\_24\_1\\ conv-3\_24\_1\end{tabular}\right]$ \\
                              &                                                                                       & conv-7\_176\_7                                                 \\ 
\midrule
Stage 1                       & $16 \times 16 + 1$                                                                           & \begin{tabular}[c]{@{}c@{}}
                                    $\left( 
                                        \begin{tabular}[c]{@{}c@{}}
                                            MHSA-32\_3, FFN-704
                                        \end{tabular}
                                    \right)$ \\ 
                                    $\left( 
                                        \begin{tabular}[c]{@{}c@{}}
                                            MHSA-32\_3, FFN-576
                                        \end{tabular}
                                    \right)$ \\
                                    $\left( 
                                        \begin{tabular}[c]{@{}c@{}}
                                            MHSA-32\_3, FFN-640
                                        \end{tabular}
                                    \right)$ \\ 
                                    $\left( 
                                        \begin{tabular}[c]{@{}c@{}}
                                            MHSA-32\_4, FFN-576
                                        \end{tabular}
                                    \right)$ \\
                                    $\left( 
                                        \begin{tabular}[c]{@{}c@{}}
                                            MHSA-32\_4, FFN-704
                                        \end{tabular}
                                    \right)$ \\ 
                                \end{tabular}
\\ 
\midrule[1pt]
\multirow{3}{*}{Stage 2}      & $8 \times 8 + 1$                                                                             & RSR-352                                                        \\ \cmidrule{2-3} 
                              & $8 \times 8 + 1$                                                                        & 
                              \begin{tabular}[c]{@{}c@{}}
                                    $\left( 
                                        \begin{tabular}[c]{@{}c@{}}
                                            MHSA-48\_10, FFN-1408
                                        \end{tabular}
                                    \right)$ \\ 
                                    $\left( 
                                        \begin{tabular}[c]{@{}c@{}}
                                            MHSA-48\_8, FFN-1408
                                        \end{tabular}
                                    \right)$ \\
                                    $\left( 
                                        \begin{tabular}[c]{@{}c@{}}
                                            MHSA-48\_8, FFN-1280
                                        \end{tabular}
                                    \right)$ \\ 
                                    $\left( 
                                        \begin{tabular}[c]{@{}c@{}}
                                            MHSA-48\_8, FFN-1408
                                        \end{tabular}
                                    \right)$ \\
                                    $\left( 
                                        \begin{tabular}[c]{@{}c@{}}
                                            MHSA-48\_10, FFN-1280
                                        \end{tabular}
                                    \right)$ \\
                                     $\left( 
                                        \begin{tabular}[c]{@{}c@{}}
                                            MHSA-48\_10, FFN-1024
                                        \end{tabular}
                                    \right)$ \\
                                \end{tabular} \\ 
\midrule[1pt]
\multirow{3}{*}{Stage 3}      & $4\times4+1$                                                                            & RSR-704                                                        \\ \cmidrule{2-3} 
                              & $4\times4+1$                                                                             & 
                              \begin{tabular}[c]{@{}c@{}}
                                    $\left( 
                                        \begin{tabular}[c]{@{}c@{}}
                                            MHSA-64\_10, FFN-2560
                                        \end{tabular}
                                    \right)$ \\ 
                                    $\left( 
                                        \begin{tabular}[c]{@{}c@{}}
                                            MHSA-64\_10, FFN-1792
                                        \end{tabular}
                                    \right)$ \\
                                    $\left( 
                                        \begin{tabular}[c]{@{}c@{}}
                                            MHSA-64\_10, FFN-2816
                                        \end{tabular}
                                    \right)$ \\ 
                                    $\left( 
                                        \begin{tabular}[c]{@{}c@{}}
                                            MHSA-64\_8, FFN-2816
                                        \end{tabular}
                                    \right)$ \\
                                    $\left( 
                                        \begin{tabular}[c]{@{}c@{}}
                                            MHSA-64\_8, FFN-2560
                                        \end{tabular}
                                    \right)$ \\
                                \end{tabular} \\ 
\bottomrule[1.2pt]
\end{tabular}
}
\vspace{1.5mm}
\caption{\textbf{Architecture of ViT-ResNAS-Tiny.}}
\label{tab:appendix:vit_resnas_tiny}
\end{table}

\begin{table}[ht]
\centering
\scalebox{0.9}{
\begin{tabular}{l|c|c}
\toprule[1.2pt]
                              & \multicolumn{1}{l|}{\begin{tabular}[c]{@{}l@{}}Output\\ Length\end{tabular}} & \textbf{ViT-ResNAS-Small}                                                   \\ 
\midrule[1.2pt]
\multirow{4}{*}{Tokenization} & \multirow{4}{*}{$16 \times 16 + 1$}                                                          & conv-3\_24\_2                                                  \\ 
& & $\left[ \begin{tabular}[c]{@{}c@{}}conv-3\_24\_1\\ conv-3\_24\_1\end{tabular}\right]$ \\
                              &                                                                                       & conv-7\_220\_7                                                 \\ 
\midrule
Stage 1                       & $16 \times 16 + 1$                                                                           & \begin{tabular}[c]{@{}c@{}}
                                    $\left( 
                                        \begin{tabular}[c]{@{}c@{}}
                                            MHSA-32\_5, FFN-880
                                        \end{tabular}
                                    \right)$ \\ 
                                    $\left( 
                                        \begin{tabular}[c]{@{}c@{}}
                                            MHSA-32\_5, FFN-880
                                        \end{tabular}
                                    \right)$ \\
                                    $\left( 
                                        \begin{tabular}[c]{@{}c@{}}
                                            MHSA-32\_7, FFN-800
                                        \end{tabular}
                                    \right)$ \\ 
                                    $\left( 
                                        \begin{tabular}[c]{@{}c@{}}
                                            MHSA-32\_5, FFN-720
                                        \end{tabular}
                                    \right)$ \\
                                    $\left( 
                                        \begin{tabular}[c]{@{}c@{}}
                                            MHSA-32\_5, FFN-720
                                        \end{tabular}
                                    \right)$ \\ 
                                    $\left( 
                                        \begin{tabular}[c]{@{}c@{}}
                                            MHSA-32\_5, FFN-720
                                        \end{tabular}
                                    \right)$ \\ 
                                \end{tabular}
\\ 
\midrule[1pt]
\multirow{3}{*}{Stage 2}      & $8 \times 8 + 1$                                                                             & RSR-440                                                        \\ \cmidrule{2-3} 
                              & $8 \times 8 + 1$                                                                        & 
                              \begin{tabular}[c]{@{}c@{}}
                                    $\left( 
                                        \begin{tabular}[c]{@{}c@{}}
                                            MHSA-48\_10, FFN-1760
                                        \end{tabular}
                                    \right)$ \\ 
                                    $\left( 
                                        \begin{tabular}[c]{@{}c@{}}
                                            MHSA-48\_10, FFN-1440
                                        \end{tabular}
                                    \right)$ \\
                                    $\left( 
                                        \begin{tabular}[c]{@{}c@{}}
                                            MHSA-48\_10, FFN-1920
                                        \end{tabular}
                                    \right)$ \\ 
                                    $\left( 
                                        \begin{tabular}[c]{@{}c@{}}
                                            MHSA-48\_10, FFN-1600
                                        \end{tabular}
                                    \right)$ \\
                                    $\left( 
                                        \begin{tabular}[c]{@{}c@{}}
                                            MHSA-48\_12, FFN-1600
                                        \end{tabular}
                                    \right)$ \\
                                     $\left( 
                                        \begin{tabular}[c]{@{}c@{}}
                                            MHSA-48\_12, FFN-1440
                                        \end{tabular}
                                    \right)$ \\
                                \end{tabular} \\ 
\midrule[1pt]
\multirow{3}{*}{Stage 3}      & $4\times4+1$                                                                            & RSR-880                                                       \\ \cmidrule{2-3} 
                              & $4\times4+1$                                                                             & 
                              \begin{tabular}[c]{@{}c@{}}
                                    $\left( 
                                        \begin{tabular}[c]{@{}c@{}}
                                            MHSA-64\_16, FFN-3200
                                        \end{tabular}
                                    \right)$ \\ 
                                    $\left( 
                                        \begin{tabular}[c]{@{}c@{}}
                                            MHSA-64\_12, FFN-3200
                                        \end{tabular}
                                    \right)$ \\
                                    $\left( 
                                        \begin{tabular}[c]{@{}c@{}}
                                            MHSA-64\_16, FFN-2880
                                        \end{tabular}
                                    \right)$ \\ 
                                    $\left( 
                                        \begin{tabular}[c]{@{}c@{}}
                                            MHSA-64\_12, FFN-2240
                                        \end{tabular}
                                    \right)$ \\
                                    $\left( 
                                        \begin{tabular}[c]{@{}c@{}}
                                            MHSA-64\_14, FFN-2560
                                        \end{tabular}
                                    \right)$ \\
                                \end{tabular} \\ 
\bottomrule[1.2pt]
\end{tabular}
}
\vspace{1.5mm}
\caption{\textbf{Architecture of ViT-ResNAS-Small.}}
\label{tab:appendix:vit_resnas_small}
\end{table}

\begin{table}[ht]
\centering
\scalebox{0.9}{
\begin{tabular}{l|c|c}
\toprule[1.2pt]
                              & \multicolumn{1}{l|}{\begin{tabular}[c]{@{}l@{}}Output\\ Length\end{tabular}} & \textbf{ViT-ResNAS-Medium}                                                   \\ 
\midrule[1.2pt]
\multirow{4}{*}{Tokenization} & \multirow{4}{*}{$16 \times 16 + 1$}                                                          & conv-3\_24\_2                                                  \\ 
& & $\left[ \begin{tabular}[c]{@{}c@{}}conv-3\_24\_1\\ conv-3\_24\_1\end{tabular}\right]$ \\
                              &                                                                                       & conv-7\_240\_7                                                 \\ 
\midrule
Stage 1                       & $16 \times 16 + 1$                                                                           & \begin{tabular}[c]{@{}c@{}}
                                    $\left( 
                                        \begin{tabular}[c]{@{}c@{}}
                                            MHSA-32\_7, FFN-960
                                        \end{tabular}
                                    \right)$ \\ 
                                    $\left( 
                                        \begin{tabular}[c]{@{}c@{}}
                                            MHSA-32\_6, FFN-960
                                        \end{tabular}
                                    \right)$ \\
                                    $\left( 
                                        \begin{tabular}[c]{@{}c@{}}
                                            MHSA-32\_7, FFN-800
                                        \end{tabular}
                                    \right)$ \\ 
                                    $\left( 
                                        \begin{tabular}[c]{@{}c@{}}
                                            MHSA-32\_8, FFN-960
                                        \end{tabular}
                                    \right)$ \\
                                    $\left( 
                                        \begin{tabular}[c]{@{}c@{}}
                                            MHSA-32\_7, FFN-880
                                        \end{tabular}
                                    \right)$ \\ 
                                    $\left( 
                                        \begin{tabular}[c]{@{}c@{}}
                                            MHSA-32\_8, FFN-880
                                        \end{tabular}
                                    \right)$ \\ 
                                    $\left( 
                                        \begin{tabular}[c]{@{}c@{}}
                                            MHSA-32\_6, FFN-800
                                        \end{tabular}
                                    \right)$ \\ 
                                \end{tabular}
\\ 
\midrule[1pt]
\multirow{3}{*}{Stage 2}      & $8 \times 8 + 1$                                                                             & RSR-640                                                        \\ \cmidrule{2-3} 
                              & $8 \times 8 + 1$                                                                        & 
                              \begin{tabular}[c]{@{}c@{}}
                                    $\left( 
                                        \begin{tabular}[c]{@{}c@{}}
                                            MHSA-48\_10, FFN-1120
                                        \end{tabular}
                                    \right)$ \\ 
                                    $\left( 
                                        \begin{tabular}[c]{@{}c@{}}
                                            MHSA-48\_14, FFN-1760
                                        \end{tabular}
                                    \right)$ \\
                                    $\left( 
                                        \begin{tabular}[c]{@{}c@{}}
                                            MHSA-48\_14, FFN-1920
                                        \end{tabular}
                                    \right)$ \\ 
                                    $\left( 
                                        \begin{tabular}[c]{@{}c@{}}
                                            MHSA-48\_16, FFN-1760
                                        \end{tabular}
                                    \right)$ \\
                                    $\left( 
                                        \begin{tabular}[c]{@{}c@{}}
                                            MHSA-48\_14, FFN-1440
                                        \end{tabular}
                                    \right)$ \\
                                     $\left( 
                                        \begin{tabular}[c]{@{}c@{}}
                                            MHSA-48\_16, FFN-1760
                                        \end{tabular}
                                    \right)$ \\
                                    $\left( 
                                        \begin{tabular}[c]{@{}c@{}}
                                            MHSA-48\_16, FFN-1920
                                        \end{tabular}
                                    \right)$ \\
                                \end{tabular} \\ 
\midrule[1pt]
\multirow{3}{*}{Stage 3}      & $4\times4+1$                                                                            & RSR-880                                                       \\ \cmidrule{2-3} 
                              & $4\times4+1$                                                                             & 
                              \begin{tabular}[c]{@{}c@{}}
                                    $\left( 
                                        \begin{tabular}[c]{@{}c@{}}
                                            MHSA-64\_16, FFN-3200
                                        \end{tabular}
                                    \right)$ \\ 
                                    $\left( 
                                        \begin{tabular}[c]{@{}c@{}}
                                            MHSA-64\_10, FFN-3840
                                        \end{tabular}
                                    \right)$ \\
                                    $\left( 
                                        \begin{tabular}[c]{@{}c@{}}
                                            MHSA-64\_16, FFN-3840
                                        \end{tabular}
                                    \right)$ \\ 
                                    $\left( 
                                        \begin{tabular}[c]{@{}c@{}}
                                            MHSA-64\_12, FFN-3200
                                        \end{tabular}
                                    \right)$ \\
                                    $\left( 
                                        \begin{tabular}[c]{@{}c@{}}
                                            MHSA-64\_16, FFN-3520
                                        \end{tabular}
                                    \right)$ \\
                                    $\left( 
                                        \begin{tabular}[c]{@{}c@{}}
                                            MHSA-64\_14, FFN-3520
                                        \end{tabular}
                                    \right)$ \\
                                \end{tabular} \\ 
\bottomrule[1.2pt]
\end{tabular}
}
\vspace{1.5mm}
\caption{\textbf{Architecture of ViT-ResNAS-Medium.}}
\label{tab:appendix:vit_resnas_medium}
\end{table}

\end{document}